\documentclass{article}
\usepackage{spconf,amsmath,graphicx}

\usepackage{xcolor}
\usepackage{multirow}
\usepackage{float}
\usepackage{cite}
\usepackage{amssymb}
\usepackage{algorithm}  
\usepackage{algpseudocode}  
\usepackage{url}

\usepackage{booktabs,multirow}
\usepackage{adjustbox}

\title{Bridging Speech and Textual Pre-trained Models with Unsupervised ASR}
\name{
\begin{tabular}{c}
Jiatong Shi$^{1}$\sthanks{This work was supported by JSALT 2022 at JHU, with gift funds from Amazon, Microsoft, and Google. The analysis and all work described in this paper were performed by the authors at CMU, NTU, and JHU. Ann Lee served as one of the advisors to the project.}, Chan-Jan Hsu$^{2}$, Holam Chung$^{2}$, Dongji Gao$^{3}$, Paola Garcia$^{3}$, \\
Shinji Watanabe$^{1}$, Ann Lee$^{4}$, Hung-yi Lee$^{2}$
\end{tabular}
}
\address{
    $^{1}$ Carnegie Mellon University, $^{2}$ National Taiwan University, $^{3}$ Johns Hopkins University, $^{4}$ Meta AI \\
    \small{\texttt{\{jiatongs,swatanb\}@cs.cmu.edu},
    \texttt{\{r09946011,r10921105,hungyilee\}@ntu.edu.tw},
    \texttt{\{dgao5,lgarci27\}@jhu.edu}}
}

\begin{document}
\ninept
\maketitle

\begin{abstract}
Spoken language understanding (SLU) is a task aiming to extract high-level semantics from spoken utterances. Previous works have investigated the use of speech self-supervised models and textual pre-trained models, which have shown reasonable improvements to various SLU tasks. However, because of the mismatched modalities between speech signals and text tokens, previous methods usually need complex designs of the frameworks. This work proposes a simple yet efficient unsupervised paradigm that connects speech and textual pre-trained models, resulting in an unsupervised speech-to-semantic pre-trained model for various tasks in SLU. To be specific, we propose to use unsupervised automatic speech recognition (ASR) as a connector that bridges different modalities used in speech and textual pre-trained models. Our experiments show that unsupervised ASR itself can improve the representations from speech self-supervised models. More importantly, it is shown as an efficient connector between speech and textual pre-trained models, improving the performances of five different SLU tasks. Notably, on spoken question answering, we reach the state-of-the-art result over the challenging NMSQA benchmark.
\end{abstract}

\section{Introduction}
\label{sec: intro}

Transfer learning has been known as one of the major methodologies in the deep learning era. It has demonstrated its success in speech and natural language \cite{bert, baevski2020wav2vec}. In transfer learning, self-supervised learning (SSL) has been a large branch of studies that focuses on leveraging large amounts of unlabelled data. According to previous literature, SSL can learn to extract contextual representations that greatly improve the performances of various downstream tasks \cite{yang2021superb}. 

Many real-world tasks involve transforming one modality to another (e.g., image caption, speech recognition/understanding, songwriting, etc.) \cite{xu2015show, yu2016automatic, qian2022training}. Usually, there is a need to understand both modalities in those tasks. Given the advances in pre-trained models, several previous works have shown significant improvements by simply applying the pre-trained models \cite{gomez2017self, baevski2019effectiveness, chang2021exploration}. In their applications, pre-trained models over source modalities are easier to adopt. 

Spoken language understanding (SLU) is an example of modalities' transfer, which aims to infer the semantic meaning from spoken utterances. 
Conventional SLU methods usually adopt a pipeline that consists of an automatic speech recognition (ASR) module and a natural language understanding (NLU) module \cite{bastianelli2020slurp}. However, in recent days, more researchers have started to focus on end-to-end modeling as it could avoid potential error propagation between the two modules. Because end-to-end SLU involves both NLP and speech processing knowledge, previous studies also revealed that SSL in each modality could improve end-to-end SLU performances. Some works have investigated the joint pre-training with both speech and text modalities \cite{bapna2021slam, bapna2022mslam}. While for most other works, researchers adopted speech SSL as a feature extractor, shown to receive improvements in both performance and efficiency \cite{yang2021superb, dinarelli2022toward, borgholt2021we, chang2022exploration, translation, arora2022espnet}. On the other hand, textual SSLs are difficult to integrate into end-to-end SLU because of the mismatched modalities. Therefore, existing works usually need specific designs to utilize those textual pre-trained models, such as deliberation modules \cite{arora2022two, chung2021splat, huang2020leveraging}, two-stage decoding \cite{lai2021semi}, K-means clustering for ``tokenization" \cite{lin2022dual}. All approaches could get decent performances on some SLU downstream tasks. However, most of them complicate the training/inference procedure and need further fine-tuning with supervision when applied to downstream tasks.

To achieve a simple way of connecting SSLs between two modalities for SLU, this paper proposes a novel usage of unsupervised ASR (UASR) to bridge SSLs between two modalities in a fully \textbf{unsupervised} manner. We start with a base framework that applies UASR to augment the speech SSL model by considering unpaired textual information. Following that, we introduce the method that uses UASR to bridge speech and textual pre-trained models.
In this work, we focus on connecting wav2vec2 \cite{baevski2020wav2vec} and a phoneme variant of ByT5 \cite{phonemet5} with wav2vec-U 2.0 \cite{liu2022towards}. 
Our experiments validate the advantages of our proposed methods in various tasks and settings. Notably, we also reach the state-of-the-art performance on the challenging NMSQA benchmark on spoken question answering \cite{lin2022dual}. 

\begin{figure*}[t]
  \centering
  \vspace{-20pt}
  \includegraphics[width=0.95\linewidth]{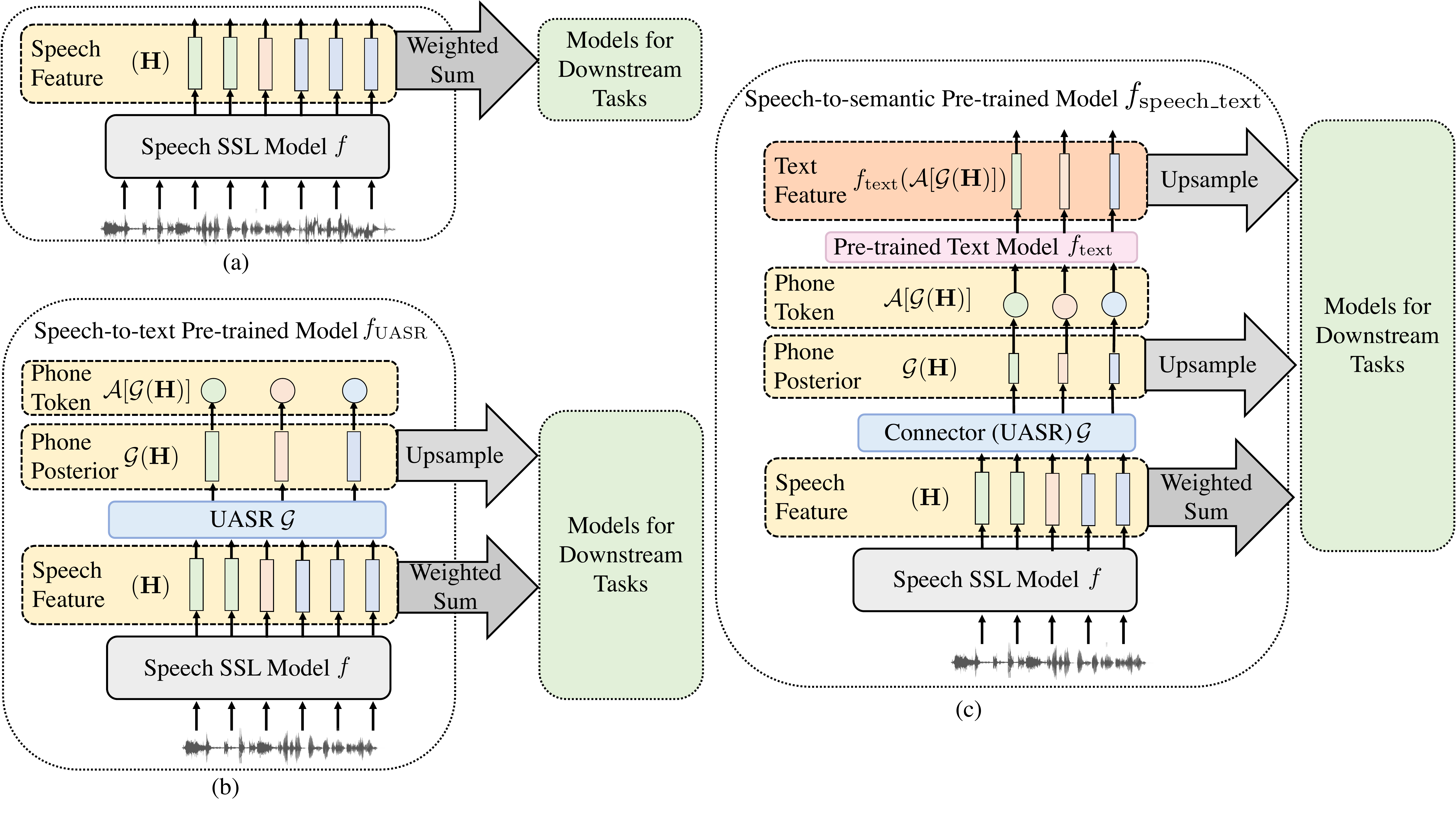}
  \vspace{-10pt}
  \caption{The comparison between different frameworks: (a) the application of speech SSL models as a feature extractor to downstream tasks, discussed in Sec~\ref{sec: background}. (b) the framework of using UASR as an augmentation module to the speech SSL, introduced in Sec~\ref{ssec: uasr-ssl}. (c) the framework of using UASR as a connector to utilize a pre-trained text model for downstream tasks, proposed in Sec~\ref{ssec: uasr-connector}.}
  \vspace{-15pt}
  \label{fig:framework}
\end{figure*}

\section{Methodology}

\subsection{Background}
\label{sec: background}

\noindent \textbf{SSL as a feature extractor}: The SSL task is to find better speech representation for downstream tasks. Previous SSL-related works usually utilize a speech-only corpus $\mathcal{D}_{\text{speech}}$. The encoder of the SSL model (i.e., $f(\cdot)$) is primarily used, after training with designed self-supervised objectives. Given a speech signal $\mathbf{X} \in \mathcal{D}_{\text{speech}}$, the encoder $f(\cdot)$ generates a hidden representation $\mathbf{H} = (h_1, h_2, ..., h_T)$, where $T$ is the number of frames of the representation. As illustrated in Fig.~\ref{fig:framework}~(a), representation $\mathbf{H}$ is used as features for an additional model, when applying to downstream tasks.

\noindent \textbf{UASR}: Unsupervised ASR focuses on utilizing unparallel speech and text to realize speech recognition systems. Before the application of SSLs, previous methods have investigated the direction by learning acoustic-text alignment, adversarial networks, and quantized auto-encoding \cite{chen2019completely, yeh2018unsupervised, liu2020towards}.

\noindent \textbf{Wav2vec-U 2.0}: Extended from its previous version \cite{baevski2021unsupervised}, Wav2vec-U 2.0 is the state-of-the-art model in UASR, which is trained in an end-to-end manner \cite{liu2022towards}. In the method, SSL (i.e., wav2vec2) is employed as a feature extractor to extract hidden representation $\mathbf{H}$. The later setup follows generative adversarial training. The generator $\mathcal{G}$ takes the segmented acoustic features to generate pseudo phonetic sequences, while the discriminator $\mathcal{C}$ focuses on distinguishing the pseudo phonetic sequence from generator $\mathcal{G}(\mathbf{H})$ and the real phonetic sequence $\mathbf{Y}^u \in \mathcal{D}_{\text{text}}$. Noted that $\mathcal{D}_{\text{text}}$ is a text-only corpus, which is unpaired with $\mathcal{D}_{\text{speech}}$. Apart from the adversarial loss, wav2vec-U 2.0 also utilizes several other types of losses (i.e., gradient penalty $\mathcal{L}_{\text{gp}}$, segment smoothness $\mathcal{L}_{\text{sp}}$, phoneme diversity $\mathcal{L}_{\text{pd}}$, and auxiliary K-means clusters $\mathcal{L}_{\text{ss}}$ from Mel frequency cepstral coefficients (MFCC)) to stabilize the training. The final optimization target is formulated as:

\begin{equation}
\begin{split}
    \label{eq: wav2vec-u2}
    \min_{\mathcal{G}} \max_{\mathcal{C}} & \mathbb{E}_{\mathbf{Y}^u} [\log \mathcal{C}(\mathbf{Y}^u)] - \mathbb{E}_{\mathbf{H}} [\log(1 - \mathcal{C(\mathcal{G}(\mathbf{H}))})] \\
    & + \lambda \mathcal{L}_{\text{gp}} + \gamma \mathcal{L}_{\text{sp}} + \eta \mathcal{L}_{\text{pd}} + \delta \mathcal{L}_{\text{ss}},
\end{split}
\end{equation}
where $\lambda, \gamma, \eta, \delta$ are the weights for losses. 

UASR with SSL provides the foundation of our following investigation. In later sections, we primarily focus on a simplified version of wav2vec-U 2.0 that is without auxiliary cluster prediction (i.e., remove $\mathcal{L}_{\text{ss}}$), as we empirically find the training is harder to converge with the proposed version in \cite{liu2022towards}.

\subsection{Problem Formulation}
\label{ssec: problem}

As speech SSLs train on a large number of speech signals from $\mathcal{D}_{\text{speech}}$, it has shown impressive performances in several speech tasks \cite{yang2021superb, tsai2022superb}. However, recent works also reveal that speech SSLs cannot handle some high-level semantic tasks (e.g., spoken question answering) \cite{lin2022dual}. To reach reasonable performances, the framework needs the following textual pre-trained model $f_{\text{text}}$ that trains on text-only corpus $\mathcal{D}_{\text{text}}$. To match the embeddings in textual pre-trained models, previous work applied K-means clustering to convert the speech SSL features into cluster IDs, which are further used as a token ID for the textual model \cite{lin2022dual}. The sequential application of speech and textual pre-trained models results in an issue of modalities mismatch between speech and text token IDs, so a fine-tuning stage is necessary to connect these modalities. On the other hand, our method can connect both SSL models in the pre-training stage without fine-tuning thanks to UASR, and build an \textbf{unsupervised speech-to-semantic pre-trained model} straightforwardly.


In the following subsections, we first add UASR as an augmentation module to the speech SSL model as a base framework, resulting in a speech-to-text pre-trained model. Then, we further extend it by combining a textual pre-trained model, where we utilize UASR as a connector and contract a speech-to-semantic pre-trained model.


\subsection{UASR as an Augmentation Module}
\label{ssec: uasr-ssl}
In this subsection, we propose using UASR as an augmentation module for the speech SSL model for the following reasons. 

\noindent \textbf{Unsupervised property}: As discussed in Sec~\ref{sec: background}, the UASR are jointly trained with inputs that include both $\mathbf{X} \in \mathcal{D}_{\text{speech}}$ and $\mathbf{Y}^u \in \mathcal{D}_{\text{text}}$. As discussed in Sec.~\ref{sec: background}, adding an unsupervised ASR network maintains the unsupervised property of the whole model, which keeps the benefits of learning from a vast amount of unsupervised data.

\noindent \textbf{Textual benefits}: The incorporation of text-only corpus $\mathcal{D}_{\text{text}}$ is the major difference between speech SSL and the proposed method. According to (\ref{eq: wav2vec-u2}), the textual information $Y^u$ is explicitly applied with the adversarial objective. 

\noindent \textbf{Performance stability}: To keep the performance stable for various tasks, we can always apply the original speech SSL with the UASR representation.

\noindent The application of UASR follows the illustration of Fig.~\ref{fig:framework}~(b). In short, the augmented model is defined as follows,
\begin{equation}
    \label{eq: uasr_ssl}
    f_{\text{UASR}}(\mathbf{H}) \triangleq (\mathbf{H}, \text{UP}(\mathcal{G}(\mathbf{H}))),
\end{equation}
where $\text{UP}(\cdot)$ is an upsampling function to match the resolution of $\mathbf{H}$ and $\mathcal{G}(\mathbf{H})$.\footnote{In our experiments, we simply use the repeat upsampling.} The $f_{\text{UASR}}(\mathbf{H})$ is used as the input of downstream models for various tasks.

\subsection{UASR as a Connector}
\label{ssec: uasr-connector}
As discussed in Sec.~\ref{sec: intro} and Sec.~\ref{ssec: problem}, the modalities mismatch issue blocks constructing pre-trained speech-to-semantic models in an unsupervised manner. Since UASR targets conducting recognition in an unsupervised fashion, it could naturally be a glue to connect the acoustic and textual pre-trained models. We argue that using UASR as a connector would have three more benefits, apart from the mitigation of the modalities mismatch issue.

\noindent \textbf{Unsupervised property}: Similar to $f_{\text{UASR}}$ in Sec.~\ref{ssec: uasr-ssl}, adding UASR as a connector can still maintain the unsupervised property of the whole model.

\noindent \textbf{Enhanced textual benefits}: As explained in Sec.~\ref{ssec: uasr-ssl}, the UASR module can introduce implicit textual knowledge from the adversarial objective. On the other hand, when applying UASR as a connector, it can adopt explicit textual resources that hold semantic information from the pre-trained text model $f_{\text{text}}$.


\noindent \textbf{Fine-tuning free}: As previous methods usually need fine-tuning to mitigate the modalities mismatch between speech and text modalities, using UASR as a connector can relax the requirement for fine-tuning since the output of UASR is a text (phoneme)-level distribution.

\noindent The application of the UASR connector follows Fig.~\ref{fig:framework}~(c). The augmented model is defined in a similar manner to $f_{\text{UASR}}$, as follows,
\begin{equation}
    \label{eq: uasr_connector}
    f_{\text{speech\_text}}(\mathbf{H}) \triangleq (\mathbf{H}, \text{UP}'(\mathcal{G}(\mathbf{H})), \text{UP}'(f_{\text{text}}(\mathcal{A}[\mathcal{G}(\mathbf{H})]))),
\end{equation}
where $\text{UP}'(\cdot)$ is another upsampling function to match the resolution of $\mathbf{H}$ and $\mathcal{A}$ is an assignment procedure that assigns phone tokens to each frame, given a phonetic posteriorgram. $\mathcal{A}$ can be a $\mathrm{GumbleSoftmax}$ module to keep the differentiable property, but it can also be an argmax operation for simplicity. Similar to Sec.~\ref{ssec: uasr-ssl}, the $f_{\text{speech\_text}}(\mathbf{H})$ is used as the input of downstream models.

\section{Experiments}

\subsection{Downstream Tasks and Datasets}

To fully investigate the proposed methods, we conduct experiments in two folds: the first freezes the parameters of the pre-trained model; the second fine-tunes the textual pre-trained model.

\noindent \textbf{Freeze setting}: In the freeze setting, we follow the exact setting of SUPERB benchmark \cite{yang2021superb}. As discussed in the previous section, we primarily focus on understanding tasks, including intent classification (IC), emotion recognition (ER), and speech translation (ST). However, as UASR are designed for recognition tasks, we also conduct experiments in ASR and phone recognition (PR). Meanwhile, we carry out the speaker identification (SID) task to verify if the UASR and the pre-trained text model are effective for semantic-related tasks.

\noindent \textbf{Fine-tuning setting}: In the fine-tuning setting, we only conduct experiments that fine-tunes the textual model to match the configuration in \cite{lin2022dual} and to avoid joint training of a large model. We adopt ESPnet \cite{inaguma2020espnet, arora2022espnet} that can support stronger downstream models. We intentionally select some different datasets that are larger than the ones in the SUPERB benchmark, evaluating the proposed method apart from the potential over-fitting issue in low-resource scenarios. Similar to the freeze setting, we focus on understanding tasks, including IC, ER, ST, and slot-filling (SF). We adopt SLURP \cite{bastianelli2020slurp} for IC and ST, IEMOCAP for ER \cite{busso2008iemocap}, and CoVOST2 for ST \cite{wang2021covost}. In addition, We also tested on a publicly available spoken question answering (SQA) NMSQA dataset\cite{lin2022dual}. 
Three test sets included are derived from SQuAD\cite{rajpurkar2016squad}, NEWSQA \cite{trischler2016newsqa} and QuAC \cite{choi2018quac}, whereas the latter two are reported together as out-of-domain (OOD) samples.

\begin{table}
\centering
\caption{\label{tab: freeze exp} Freeze setting experiments with UASR and PT5: pre-trained models are frozen during the training. The ``+UASR" option utilizes the method explained in Sec.~\ref{ssec: uasr-ssl}, while the ``+UASR +PT5" option utilizes the method showed in Sec.~\ref{ssec: uasr-connector}. }
\begin{tabular}{l|ccccc}
\toprule 
SSL & PR($\downarrow$) & ASR($\downarrow$) & IC($\uparrow$) & ST($\uparrow$) & SID($\uparrow$) \\ 
\midrule
wav2vec2 & 5.51 & 3.79 & 94.38 & 13.01 & \textbf{83.69}\\
\quad +UASR & \textbf{4.53} & \textbf{3.76} & 94.33 & 13.00 & 82.85 \\
\quad +UASR +PT5 & 4.68 & 3.97 & \textbf{94.88} & \textbf{13.53} & 82.74 \\
\bottomrule
\end{tabular}
\vspace{-15pt}
\end{table}

\begin{table*}
\centering
\caption{\label{tab: fine-tune exp} Fine-tuning setting experiments with UASR and PT5: textual pre-trained models are jointly fine-tuned with downstream models during training. \textbf{B} utilizes the method explained in Sec.~\ref{ssec: uasr-ssl}, while \textbf{D} utilize the method showed in Sec.~\ref{ssec: uasr-connector}. Others models (i.e., \textbf{A} and \textbf{C}) and more detailed configurations are introduced in Sec.~\ref{ssec: experimental setings}.}
\begin{tabular}{l|cc|cccc}
\toprule 
 & Connector & Text Model & IC($\uparrow$) & SF($\uparrow$) & ER($\uparrow$) & ST($\uparrow$) \\ 
\midrule
\textbf{A} & /\ & /\ & 82.82 & 65.82 & 66.35 & 22.1 \\
\textbf{B} & UASR & /\ & 82.93 & 64.93 & 64.11 & 22.2 \\
\textbf{C} & K-means & PT5 & 86.41 & 71.70 & 72.61 & 22.5 \\
\textbf{D} & UASR & PT5 & \textbf{87.10} & \textbf{74.03} & \textbf{73.57} & \textbf{24.3} \\
\bottomrule
\end{tabular}
\vspace{-15pt}

\end{table*}

\begin{table}
\centering
\caption{\label{tab: ablation exp} Effect of different textual pre-trained models when using UASR as a connector: \textbf{E} employs a T5 architecture with randomized parameters; \textbf{F} uses the original ByteT5; \textbf{D} applies the phoneme T5 introduced in Sec~\ref{ssec: experimental setings}.}
\begin{tabular}{l|c|cccccc}
\toprule 
 & Text Model & IC($\uparrow$) & SF($\uparrow$) & ER($\uparrow$) & ST($\uparrow$) \\ 
\midrule
\textbf{E} & RT5 & 86.61 & 71.83 & 72.19 & 23.9  \\
\textbf{F} & TT5 & 85.93 & 72.31 & 72.19 & \textbf{24.6}  \\
\midrule
\textbf{D} & PT5 & \textbf{87.10} & \textbf{74.03} & \textbf{73.57} & 24.3  \\
\bottomrule
\end{tabular}
\end{table}

\begin{table}
\centering
\caption{\label{tab: sqa exp}  Effect of different textual pre-trained models for SQA when using UASR as a connector. The Longformer model is the state-of-the-art model in \cite{lin2022dual}.}
\begin{tabular}{l|c|cc|cc}
\toprule 
& \multirow{2}{*}{Text Model}  & \multicolumn{2}{c|}{SQuAD}& \multicolumn{2}{|c}{OOD}  \\ 

& & AOS($\uparrow$) & FF1($\uparrow$) & AOS($\uparrow$) & FF1($\uparrow$) \\
\midrule
\cite{lin2022dual} & Longformer & 49.1 & 55.9 & / & /
\\
\textbf{F} & TT5 & 56.0 & 61.5 & 38.4 & 42.4 \\
\midrule
\textbf{D} & PT5 & \textbf{65.8} & \textbf{69.7} & \textbf{42.1} & \textbf{46.2} \\
\bottomrule
\end{tabular}
\vspace{-15pt}
\end{table}


\subsection{Experimental Settings}
\label{ssec: experimental setings}

\noindent
\textbf{Speech SSL}: Given that the UASR model is based on wav2vec2 \cite{baevski2020wav2vec}, we adopt pre-trained wav2vec2 that trained on LibriLight \cite{kahn2020libri}.

\noindent
\textbf{UASR}: To keep the same as \cite{liu2022towards}, we utilize Librispeech speech data for the speech-only corpus $\mathcal{D}_{\text{speech}}$ and Librispeech language modeling data for the text-only corpus $\mathcal{D}_{\text{text}}$. As noted in Sec~\ref{ssec: problem}, we do not use $\mathcal{L}_{\text{ss}}$. For other settings, we use the hyper-parameters released on Fairseq.\footnote{\scriptsize{\url{https://github.com/facebookresearch/fairseq/blob/main/examples/wav2vec/unsupervised/config/gan/w2vu2.yaml}}}

\noindent \textbf{Pre-trained text model}: 
As discussed in Sec.~\ref{sec: intro} and Sec.~\ref{ssec: uasr-connector}, we select PhonemeT5 (PT5)\footnote{\scriptsize{\url{https://github.com/voidful/t5lephone}}} \cite{phonemet5} as our main textual pre-trained model, because it has the same token space as our UASR model. PT5 is a variant of ByT5 \cite{xue2022byt5}, which uses phonemicized text\footnote{Each phoneme token is mapped to a single character with a predefined mapping dictionary to reduce the sequence length.} as input with the span reconstruction objective.

\noindent \textbf{Model candidates}: For the freeze setting, aligned with Fig.~\ref{fig:framework}, we compare the models in three settings, including speech pre-trained model (i.e., wav2vec2), speech pre-trained model with UASR, and speech pre-trained model with UASR + textual pre-trained model (i.e., PT5). For the fine-tuning setting, we compare four settings: \textbf{A} utilizes only the speech pre-trained model; \textbf{B} applies UASR as an augmented module to the speech pre-trained model; \textbf{C} employs K-means as a connector between speech and textual pre-trained models \cite{lin2022dual};\footnote{We adopt the official K-means checkpoint at \scriptsize{\url{https://dl.fbaipublicfiles.com/textless_nlp/gslm/w2v2/km50/km.bin}}. 50-cluster version is selected because UASR has a similar vocabulary size (i.e., 45).} \textbf{D} uses UASR instead of K-means as a connector. 

To further investigate the effects of different pre-trained text models, we provide two other models with the same network architecture, namely randomized T5 (RT5) and textual T5 (TT5). RT5 is a model with the same architecture as PT5 but with randomized parameters. TT5 is the original version of ByT5 that is pre-trained on text data \cite{xue2022byt5}. Because of the token mismatch (i.e., phone tokens from UASR and byte-level tokens from ByT5), we utilize a randomly initialized dictionary to map UASR's phone tokens into bytes, similar to the K-means approach \cite{lin2022dual}. Given the same framework as \textbf{D}, we name \textbf{E} as the one with RT5 and \textbf{F} as the one with TT5. Except for SQA which utilizes whole pre-trained models (i.e., both encoder and decoder), other tasks only employ the encoders of the T5 variants (i.e., \textbf{D}, \textbf{E}, and \textbf{F}). 

\noindent
\textbf{Dowstream models}: For the freeze setting, we follow exact settings as the SUPERB public benchmark, which adopts simple downstream models (e.g., stacked linear or recurrent layers). For the fine-tuning setting, we utilize ESPnet \cite{arora2022espnet, inaguma2020espnet} for the downstream models except for SQA. For IC, SF, and ER, we adopt a conformer-based encoder-decoder network with the hybrid connectionist temporal classification (CTC)/attention style of training \cite{guo2021recent, watanabe2017hybrid}. While for ST, we adopt a transformer-based encoder-decoder with auxiliary ASR loss computed from CTC. Detailed network settings follow the corresponding configuration.\footnote{\scriptsize{\url{https://github.com/espnet/espnet/blob/master/egs2/{slurp_entity, slurp, iemocap, covost1}/{asr1, st1}}}} For classification tasks like IC and ER, we also include ASR transcriptions as a joint learning objective. For all models in ESPnet, we apply specaugment to the hidden states from pre-trained models \cite{yang2021superb}. For SQA, it is difficult to hold the whole speech SSL features for the spoken context of answers. Therefore, we employ a two-step approach. Specifically, questions and contexts are converted into time-coded phoneme sequences by UASR. Given questions and contexts, the PT5 is fine-tuned to predict answer time spans in the context. The model is trained for 3 epochs with a 3e-4 learning rate and the best model is selected based on the performances on the development set.

\noindent
\textbf{Evaluation metric}: For PR, we adopt phone error rate (PER), while for ASR, we utilize word error rate (WER). For IC and ER, we employ the accuracy rate. For slot-filling, we apply the slot-type F1 score. For ST, we use the BLEU score. For SQA,  we adopt the same evaluation metrics as \cite{lin2022dual}, including the area overlapping score (AOS) and frame F1 (FF1). Note that when the answers cannot be extracted from the given context, we set AOS and FF1 to 0.


\subsection{Results and Discussion}
We first discuss the results of the freeze setting as shown in Table~\ref{tab: freeze exp}. As UASR is optimized for PR, it is reasonable to have better performances in PR. However, ASR does not have significant gain with UASR, which shows that the PR advantage from UASR may not directly enhance the speech SSL. When it comes to understanding tasks (i.e., IC and ST), the one with UASR and PT5 achieves the best performance, demonstrating the semantic benefits introduced by PT5. Given the wav2vec2, UASR, and PT5 do not update their parameters, we could infer that the UASR as a connector could mitigate the modalities mismatch issue discussed in Sec.~\ref{ssec: problem}. Last, it is reasonable that UASR and PT5 cannot improve the performances of speaker identification, as their features barely contain speaker-related information at the design level.

Table~\ref{tab: fine-tune exp} shows the results in the fine-tuning setting. For all the selected tasks, \textbf{D} achieves the best performance by using UASR as a connector. When applying K-means as a connector (i.e., \textbf{C}), we find similar performance improvements in all four tasks as \cite{lin2022dual}. However, with UASR as a connector, the improvements are even larger.

As in the ablation study shown in Table~\ref{tab: ablation exp}, the one with PT5~(i.e., \textbf{D}) reaches the best performance for IC, SF, and ER, while the one with TT5~(i.e.,\textbf{F}) has the best performance for ST. Based on the experimental results, we would argue that there is a balance between modalities matching and textual semantics. 
As for text generation tasks, \textbf{F} has better results. Table~\ref{tab: sqa exp} shows our proposed method on SQA. We only conduct the model with textual pre-trained models, as we find the text information is essential to train the model. For example, model \textbf{E} can hardly converge for the task, resulting in 0.0 scores for the test sets. On the contrary, the model \textbf{D} with matched modalities has shown outstanding performances, significantly better than the model \textbf{F} and the best model in \cite{lin2022dual}.

\section{Conclusion}
Though pre-trained models in speech and languages have shown to be effective, it is difficult to sequentially apply speech and language pre-trained models to scenarios of spoken language understanding.
In this work, we propose using unsupervised mapping (i.e., UASR in our case) to mitigate the mismatch between modalities. We first show that adding UASR could enhance the original speech pre-trained model, then UASR is applied as a connector between speech and textual pre-trained models. Extensive experiments, in various tasks and settings, demonstrate the effectiveness of our proposed method, which also shows possible solutions to other scenarios other than speech and language.



\bibliographystyle{IEEEbib}
\bibliography{strings,refs}

\end{document}